\documentclass[conference]{IEEEtran}
\IEEEoverridecommandlockouts
\usepackage{cite}
\usepackage{amsmath,amssymb,amsfonts}
\usepackage{algorithmic}
\usepackage{graphicx}
\usepackage{textcomp}
\usepackage{graphicx} 
\usepackage{amsmath,amsfonts,amssymb,mathrsfs}
\usepackage[scr=rsfso]{mathalfa}
\usepackage[table,xcdraw]{xcolor}
\usepackage{bbm}
\usepackage{verbatim}
\usepackage{pgfplots}
\pgfplotsset{compat=1.18}	
\usetikzlibrary{pgfplots.statistics} 
\usetikzlibrary{matrix}
\usepgfplotslibrary{groupplots}
\usepackage{tabularx}
\usepackage{multirow}
\usepackage{url}
\usepackage{array}
\usepackage{booktabs}
\usepackage{adjustbox}
\usepackage{setspace}

\usepackage{color}

\def\BibTeX{{\rm B\kern-.05em{\sc i\kern-.025em b}\kern-.08em
    T\kern-.1667em\lower.7ex\hbox{E}\kern-.125emX}}
\begin{document}

\title{Timing Is Everything: Finding the Optimal Fusion Points in Multimodal Medical Imaging}

\author{
    \IEEEauthorblockN{Valerio Guarrasi\IEEEauthorrefmark{1},
                      Klara Mogensen\IEEEauthorrefmark{2},
                      Sara Tassinari\IEEEauthorrefmark{1},
                      Sara Qvarlander\IEEEauthorrefmark{2}, and
                      Paolo Soda\IEEEauthorrefmark{1}\IEEEauthorrefmark{2}}
    \IEEEauthorblockA{\IEEEauthorrefmark{1}Research Unit of Computer Systems and Bioinformatics, Department of Engineering, \\
    Università Campus Bio-Medico di Roma, Rome, Italy \\
    Emails: valerio.guarrasi@unicampus.it, sara5tassinari@gmail.com, p.soda@unicampus.it}
    \IEEEauthorblockA{\IEEEauthorrefmark{2}Department of Diagnostics and Intervention, Biomedical Engineering and Radiation Physics, Umeå University, Umeå, Sweden \\
    Email: klara.mogensen@umu.se, sara.qvarlander@umu.se, paolo.soda@umu.se}
}

\maketitle

\begin{abstract}
Multimodal deep learning harnesses diverse imaging modalities, such as MRI sequences, to enhance diagnostic accuracy in medical imaging. A key challenge is determining the optimal timing for integrating these modalities—specifically, identifying the network layers where fusion modules should be inserted. Current approaches often rely on manual tuning or exhaustive search, which are computationally expensive without any guarantee of converging to optimal results. We propose a sequential forward search algorithm that incrementally activates and evaluates candidate fusion modules at different layers of a multimodal network. At each step, the algorithm retrains from previously learned weights and compares validation loss to identify the best-performing configuration. This process systematically reduces the search space, enabling efficient identification of the optimal fusion timing without exhaustively testing all possible module placements. The approach is validated on two multimodal MRI datasets, each addressing different classification tasks. Our algorithm consistently identified configurations that outperformed unimodal baselines, late fusion, and a brute-force ensemble of all potential fusion placements. These architectures demonstrated superior accuracy, F-score, and specificity while maintaining competitive or improved AUC values. Furthermore, the sequential nature of the search significantly reduced computational overhead, making the optimization process more practical. By systematically determining the optimal timing to fuse imaging modalities, our method advances multimodal deep learning for medical imaging. It provides an efficient and robust framework for fusion optimization, paving the way for improved clinical decision-making and more adaptable, scalable architectures in medical AI applications.

\end{abstract}



\section{Introduction} \label{sec:Introduction}

Multimodal deep learning has emerged as a transformative paradigm in medical imaging, capitalizing on the complementary strengths of diverse imaging modalities, to enhance diagnostic accuracy and clinical decision-making.
Unlike single-modal approaches, which may suffer from incomplete or noisy information, multimodal methods integrate heterogeneous data sources to generate richer and more informative representations of anatomy and pathology.
For instance, combining different magnetic resonance imaging (MRI) sequences improves lesion localization and characterization, fostering more accurate diagnoses and personalized treatment plans~\cite{bib:huang2024multimodal, bib:pei2023review}.
Recent advances in deep neural networks have further accelerated this trend, enabling automated feature extraction and fusion at various representation levels while reducing reliance on handcrafted features and domain-specific heuristics.
These innovations have significantly improved performance in tasks such as tumor segmentation and disease classification, with applications in oncology, neurology, and cardiology~\cite{bib:sun2023scoping, bib:xu2019deep}.
Beyond technical advancements, the clinical implications of multimodal learning, e.g., earlier disease detection, personalized treatment planning, and improved patient outcomes, underscore its transformative potential in medical image analysis~\cite{bib:simonfuture, bib:artsi2024advancing}.

Despite its promise, optimizing the fusion of multiple imaging modalities within a deep learning framework remains a critical challenge.
This challenge revolves around three core questions: \textit{Which} networks best process each modality, \textit{How} to design effective fusion modules, and crucially, \textit{When} to integrate modalities within the network pipeline.
While substantial progress has been made in addressing the \textit{Which} and \textit{How} questions~\cite{bib:guarrasi2024multimodal, bib:ruffini2024multi, bib:caruso2024deep, bib:rofena2024deep, bib:guarrasi2023multi, bib:caruso2022multimodal, bib:guarrasi2022optimized, bib:mogensen2025optimized}, the \textit{When} dimension remains underexplored~\cite{bib:yao2024integrating, bib:sherwani2024systematic}.

Fusion timing strategies, answering \textit{When} the fusion should occur, in multimodal learning can be broadly categorized into early, late, and intermediate fusion.
Early fusion combines modalities at low-level feature stages but may fail to leverage the full discriminatory power of each modality.
Conversely, late fusion methods often miss critical cross-modal interactions that emerge at intermediate representation levels.
Intermediate fusion, bonded with deep networks, offers a promising middle ground, but it introduces a combinatorial explosion of potential integration points, making exhaustive evaluation computationally infeasible~\cite{bib:guarrasi2024systematic, bib:heiliger2023beyond, bib:sun2023scoping}.
Furthermore, modality-specific characteristics such as resolution, contrast, and noise distributions add complexity, as the optimal fusion point often depends on the specific task and dataset.
These challenges underscore the importance of systematically addressing the \textit{When} dimension, identifying the ideal stage at which to fuse modalities.
While existing strategies provide partial solutions, they lack a principled, data-driven approach to fusion timing, leaving a critical gap in multimodal learning for medical imaging.

To address this gap, we propose a novel Sequential Forward Search Algorithm (SFSA) to systematically identify optimal fusion points in multimodal deep learning architectures.
Our method incrementally evaluates candidate fusion modules, and halts the exploration once the performance reaches a plateau.
By advancing the state-of-the-art in multimodal medical imaging, our framework provides a scalable, efficient solution to the fusion timing problem, offering actionable insights for the development of adaptive architectures.
The primary contributions of this work are threefold:
\begin{itemize}
    \item Sequential Fusion Search Algorithm: We introduce a data-driven approach, SFSA, that incrementally activates and evaluates candidate fusion modules. This approach enables efficient discovery of optimal fusion timings without exhaustive search, reducing computational overhead.
    \item Adaptive Multimodal Integration: By integrating a Multimodal Transfer Module (MMTM)~\cite{bib:MMTM} at strategically selected layers, our framework capitalizes on complementary features from multiple modalities, leading to more robust and discriminative representations.
    \item Evaluation and Benchmarking: We validate the proposed method on two distinct publicly available multimodal MRI datasets, comparing its performance against unimodal baselines, late-fusion models, and an exhaustive fusion search. Our results consistently demonstrate improved classification metrics and reduced training time, underscoring both the effectiveness and the scalability of the approach.
\end{itemize}
These contributions hold promise for improved diagnostic accuracy, more robust clinical decision-making, and the broader adoption of multimodal learning in medical diagnostics and patient care.

The remainder of this paper is organized as follows: 
Section~\ref{sec:Related_Work} reviews the related work on multimodal fusion strategies; Section~\ref{sec:Methods} introduces our proposed methodology; Section~\ref{sec:Experimental_Setup} describes the experimental setup, including datasets, training protocols, and baseline configurations; Section~\ref{sec:Results_and_Discussions} presents the results and provides a comprehensive discussion, comparing our approach against competitor methods and analyzing its computational efficiency; Section~\ref{sec:Conclusions} concludes the paper, summarizing key findings, discussing limitations, and suggesting directions for future research.

\section{Related Work} \label{sec:Related_Work}

Multimodal deep learning integrates information from diverse sources through three primary fusion paradigms: early, late, and intermediate fusion, each offering unique advantages and limitations.
Early fusion, also known as feature-level fusion, combines raw input data or low-level features at the initial layers of a network.
While this approach can exploit shared low-level patterns from the outset, it risks diluting modality-specific features if one modality dominates or introduces noise~\cite{bib:gao2020survey, bib:hussain2024revolutionizing}.
Late fusion, on the other hand, processes each modality independently until the final prediction stage, where high-level feature representations or derived outputs are merged.
This strategy preserves modality-specific characteristics and simplifies network design by allowing each stream to be optimized independently.
However, it may fail to capture subtle cross-modal interactions that emerge at intermediate representation levels, potentially limiting performance~\cite{bib:gao2020survey, bib:ngiam2011multimodal}.
Intermediate fusion addresses these shortcomings by integrating modalities at one or more mid-level layers of the deep network, enabling the capture of complex interdependencies between modalities~\cite{bib:guarrasi2024systematic}.
This paradigm can produce richer feature representations and improve performance for tasks such as classification and segmentation. However, determining the optimal layers or stages for integration is nontrivial, often it requires extensive trial-and-error or heuristic-driven tuning. Recent advancements, such as attention mechanisms, gating functions, and learnable parameters, have introduced adaptive strategies to guide the fusion process, yet these approaches lack a principled framework for determining fusion timing~\cite{bib:xue2023dynamic, bib:baltruvsaitis2018multimodal}.

In medical imaging, these fusion paradigms have been extensively applied to tasks like multimodal tumor segmentation and disease classification~\cite{bib:hussain2024revolutionizing}.
Leveraging complementary imaging modalities has demonstrated significant potential to refine diagnostic decision-making.
Among these, intermediate fusion stands out for its ability to harness the unique strengths of each modality~\cite{bib:ngiam2011multimodal}.
However, identifying the optimal fusion points remains an open challenge, underscoring the need for systematic methods to optimize fusion configurations.

Neural architecture search (NAS) has gained traction as a systematic approach for optimizing network topologies without relying entirely on human expertise~\cite{bib:xue2023dynamic}.
Early NAS methods employed reinforcement learning and evolutionary algorithms to iteratively refine candidate architectures, achieving notable success but at the cost of prohibitively high computational overhead.
Recent advancements in NAS have introduced differentiable search spaces and gradient-based optimization, significantly reducing computational demands and enabling faster convergence to high-performing architectures.
However, these methods predominantly focus on optimizing single-modal networks, often assuming fixed operations or layers rather than addressing the unique challenges of fusing multimodal data streams.
In multimodal contexts, some efforts have incorporated NAS principles to optimize fusion strategies~\cite{bib:perez2019mfas, bib:valada2017adapnet, bib:ma2021novel}.
These approaches typically focus on selecting fusion operations or tailoring modality-specific subnetworks, rather than systematically identifying the optimal stages for integration~\cite{bib:long2024feature}.

Our SFSA directly addresses the challenge of determining optimal fusion timing within multimodal networks.
Unlike exhaustive search strategies that evaluate all possible configurations, our method incrementally activates fusion modules and evaluates their impact on performance, halting exploration when no further improvement is observed.
By leveraging previously learned weights, this selective, data-driven approach significantly reduces training time while maintaining high efficiency.
Compared to existing NAS-inspired multimodal frameworks, our algorithm introduces a more constrained yet purposeful exploration of the design space, offering a practical and scalable solution to the fusion timing problem in multimodal deep learning architectures.

\section{Methods} \label{sec:Methods}

\begin{figure*}[ht]
\centering
\includegraphics[width=\textwidth]{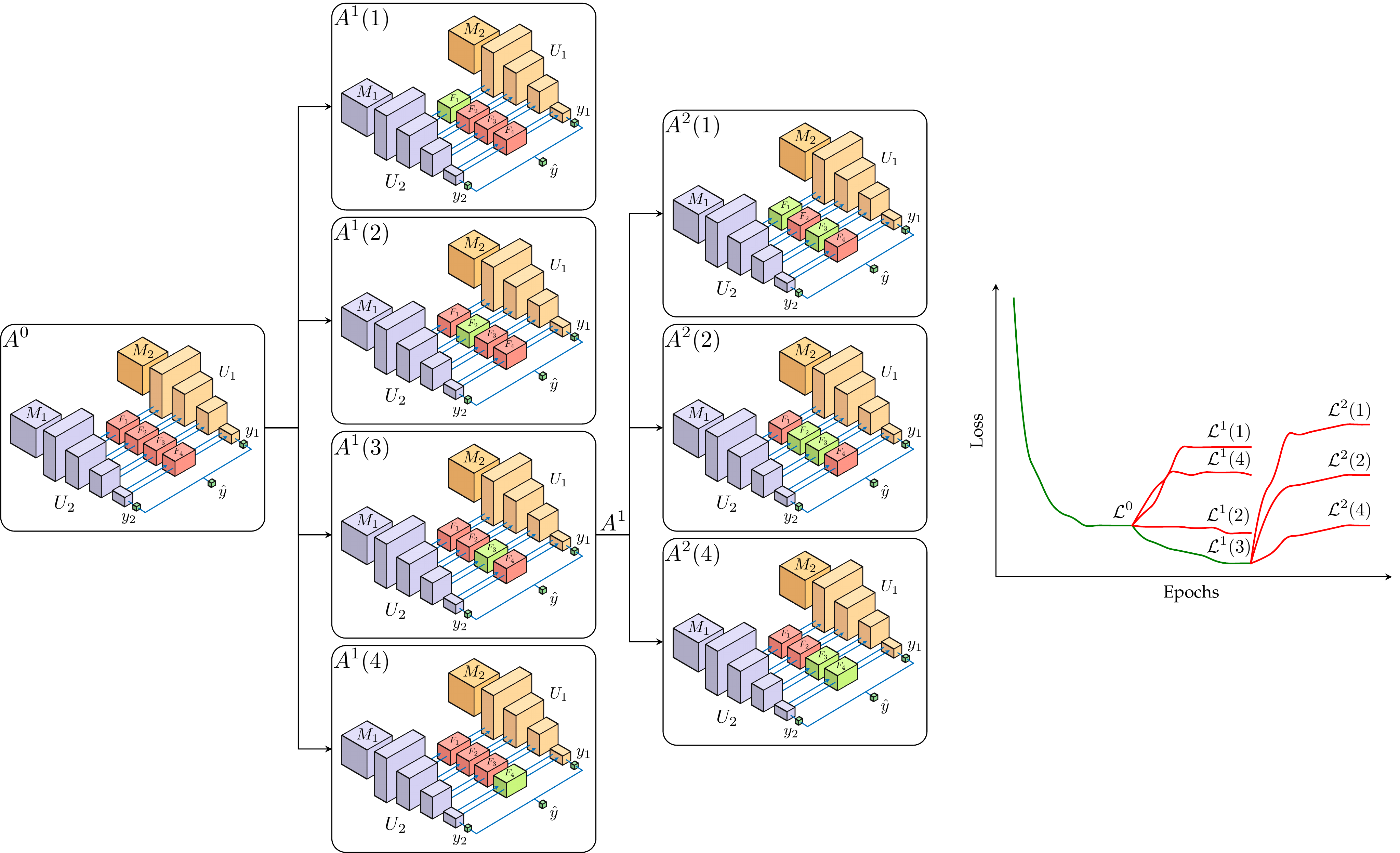}
\caption{Illustration of the SFSA. Starting from the left, the baseline configuration $A^0$ consists of the unimodal modules $U_1$ and $U_2$, which process the respective modalities $M_1$ and $M_2$ and produce the corresponding outputs $y_1$ and $y_2$, which are then merged into $\hat{y}$. These modules have four potential fusion points ($F_1$, $F_2$, $F_3$, $F_4$): the fusion modules are depicted in red when inactive and in green when active. Each single-module configuration ($A^1(1)$, $A^1(2)$, $A^1(3)$, $A^1(4)$) is evaluated individually, and the configuration yielding the greatest improvement ($A^1(3)$, with $F_3$ active in this example) is selected, as shown in the plot of the loss (i.e., $\mathcal{L}^1(3)$ has the lowest value, despite $\mathcal{L}^1(2)$ also being lower than $\mathcal{L}^0$). Attempts to add a second module on top of the best single-module configuration ($A^1$) do not result in further improvements (the loss plot shows that $\mathcal{L}^2(1)$, $\mathcal{L}^2(2)$ and $\mathcal{L}^2(4)$ have higher loss values), confirming that activating only $F_3$ provides the optimal fusion strategy for this scenario.}
\label{fig:method}
\end{figure*}

Our proposed framework consists of multiple unimodal deep networks, each specialized in processing a single imaging modality, and a set of candidate fusion modules that can be selectively activated at various intermediate layers.
By activating these fusion modules at carefully selected points, we aim to identify the optimal configuration, i.e., \textit{When} fusions should occur, that yields improved performance with minimal computational overhead.
Figure~\ref{fig:method} provides a schematic representation of the methodology, with further details elaborated in the following sections.

\subsection{Notation and Model Architecture}
Consider a set of imaging modalities $\mathcal{M} = \{M_1, M_2, \dots, M_n\}$ (e.g., different MRI sequences). For each modality $M_i$, we define a corresponding unimodal deep network $U_i(\cdot; \theta_i)$, parameterized by $\theta_i$. Given an input image modality $M_i \in \mathbb{R}^{H \times W}$, the network $U_i$ produces a hierarchy of features $\{f_{i}^{j} \mid j = 1, \dots, l\}$, where $f_{i}^{j}$ is the feature representation extracted at layer $j$.

To integrate information across modalities, we introduce a set of fusion modules $\mathcal{F} = \{F_1, F_2, \dots, F_l\}$. Each fusion module $F_j(\cdot; \phi_j)$, parameterized by $\phi_j$, operates on a set of intermediate unimodal features $\{f_{1}^{j}, f_{2}^{j}, \dots, f_{n}^{j}\}$ extracted at a specific layer $j$. The fusion module produces a joint representation $z_j$:
\begin{equation}
    z_j = F_j(f_{1}^{j}, f_{2}^{j}, \dots, f_{n}^{j}; \phi_j).
\end{equation}
This integrated representation, computed at the layer $j$, is then fed back into each unimodal pathway, allowing subsequent layers to refine their modality-specific features using cross-modal context.

At training time, each unimodal network $U_i$ outputs an estimate of the posterior classification probabilities $y_i \in \mathbb{R}^o$, where $o$ is the number of output classes.
Here we compute the final prediction $\hat{y}$ by averaging these unimodal outputs:
\begin{equation}
    \hat{y} = \frac{1}{n} \sum_{i=1}^{n} y_i.
\end{equation}
Before any fusion modules are activated, the unimodal networks can be trained independently or jointly. The activation of fusion modules integrates multimodal information progressively, influencing subsequent feature extraction stages.

\subsection{Training Objective}
We employ a combination of cross-entropy loss functions during training, with each loss derived from the corresponding unimodal deep network.
For the $i$-th unimodal output, we have:
\begin{equation}
    \mathcal{L}_i = - \sum_{k=1}^{o} y_k^* \log(y_{i,k}),
\end{equation}
where $y^* \in \{0, 1\}^o$ is the one-hot encoded ground-truth label for a given sample, and $y_{i,k}$ is the predicted probability for class $k$ from the $i$-th unimodal network. The total loss is then:
\begin{equation}
    \mathcal{L} = \frac{1}{n} \sum_{i=1}^{n} \mathcal{L}_i.
\end{equation}
This objective encourages each unimodal pathway to align its predictions with the ground truth.

When fusion modules are active, they provide multimodal context that can refine each unimodal feature set, and because all components are differentiable, gradients can be propagated through both unimodal and fusion modules. This joint optimization process adaptively enhances representations across modalities, leading to improved classification performance.

\subsection{Multimodal Transfer Module (MMTM)}
As fusion modules $F_j$ with $j = 1, \dots, l$, we use the Multimodal Transfer Module (MMTM)~\cite{bib:MMTM} as it is a key fusion component designed to enhance information exchange between modalities.
Considering intermediate feature maps $\{f_{1}^{j}, f_{2}^{j}, \dots, f_{n}^{j}\}$ at layer $j$, the MMTM transforms these unimodal features into recalibrated representations $\{\tilde{f}_{1}^{j}, \tilde{f}_{2}^{j}, \dots, \tilde{f}_{n}^{j}\}$ that emphasize discriminative patterns and suppress less relevant features.

The process consists of a \textit{compression} phase, where the modality-specific features are concatenated and are projected into a lower-dimensional space:
\begin{equation}
    z = \sigma(W_c [f_{1}^{j} \oplus f_{2}^{j} \oplus \cdots \oplus f_{n}^{j}] + b_c),
\end{equation}
where $\oplus$ denotes concatenation, $\sigma(\cdot)$ is a nonlinear activation (e.g., ReLU), and $W_c, b_c$ are learnable parameters.

Then an \textit{excitation} phase follows, which uses $z$ as a joint representation, generating modality-specific gating vectors:
\begin{equation}
    g_i = \text{softmax}(W_{e,i} z + b_{e,i}),
\end{equation}
where $W_{e,i}, b_{e,i}$ are learnable parameters for the modality $M_i$. The softmax ensures that each dimension of $g_i$ represents a relative weighting within modality $M_i$'s feature space.
Finally, a recalibration phase applies these weights to the original features:
\begin{equation}
    \tilde{f}_{i}^{j} = g_i \odot f_{i}^{j},
\end{equation}
where $\odot$ denotes element-wise multiplication. This produces refined modality-specific features that incorporate cross-modal cues, potentially improving accuracy and robustness.

\subsection{Sequential Forward Search Algorithm (SFSA)}

To determine the optimal fusion configuration, we propose an approach that incrementally refines the multimodal network architecture by adding one fusion module at a time.
Each addition is accepted only if it improves the validation loss.
This strategy efficiently navigates the search space, reducing the need for expensive exhaustive exploration.
The algorithm follows these steps:

\noindent\textit{Initialization:}  
The algorithm starts with a baseline configuration that contains no fusion modules. Specifically, let the initial configuration be:  
\begin{equation}
A^0 = \{\}, \quad \mathcal{L}^0 = \mathcal{L}(U_1, \dots, U_n, A^0, D^{\text{val}})
\end{equation}
where $D^{\text{val}}$ is the validation dataset. Here, $\mathcal{L}^0$ represents the validation loss obtained without any fusion modules.

\noindent\textit{Single-Module Exploration:}  
From the baseline configuration, the algorithm explores the potential addition of each candidate fusion module $F_j$ at position $j = 1, \ldots,  l$. For every candidate module $F_j$, a new configuration is created as follows:  
\begin{equation}
A^1(j) = \{F_j\}
\end{equation}  
The network is retrained starting from the weights obtained in $A^0$, and the resulting validation loss $\mathcal{L}^1(j) = (U_1, \dots, U_n, A^1(j), D^{\text{val}})$ is evaluated.

\noindent\textit{Best Module Selection:}  
After evaluating all candidate fusion modules, the algorithm compares the validation losses $\mathcal{L}^1(j)$ with the baseline loss $\mathcal{L}^0$.
If one or more candidate modules result in an improvement, the best-performing module $F_{j^*}$ is selected:  
\begin{equation}
\text{if } \mathcal{L}^1(j^*) < \mathcal{L}^0: \quad A^1 = A^1(j^*), \quad \mathcal{L}^1 = \mathcal{L}^1(j^*)
\end{equation}  
If no candidate module improves performance, the algorithm terminates and returns the baseline configuration $A^0$.

\noindent\textit{Iterative Expansion:}  
If the addition of a single module improves performance, the algorithm proceeds by attempting to add a second module. Starting from the best-known configuration $A^1$, each remaining candidate module $F_{j}$ at position $j \neq j^*$ is evaluated. For each candidate, a new configuration is formed as:  
\begin{equation}
A^2(j) = A^1 \cup \{F_{j}\}
\end{equation}  
The network is retrained from the weights obtained in $A^1$, and the corresponding validation loss $\mathcal{L}^2(j)$ is computed. 
If an improvement is observed with $\mathcal{L}^2(j) < \mathcal{L}^1$, the configuration and loss are updated with the best-performing candidate configuration:
\begin{equation}
A^2 = A^2(j), \quad \mathcal{L}^2 = \mathcal{L}^2(j)
\end{equation}  
Otherwise, the algorithm retains the configuration $A^1$.
This process is repeated iteratively. At each iteration $t$, the algorithm evaluates whether adding a new module improves performance. If so, the configuration is updated as:  
\begin{equation}
\text{if } \mathcal{L}^t(j) < \mathcal{L}^{t-1}: \quad A^t = A^{t-1} \cup \{F_{j}\} 
\end{equation}  
If no further improvement is observed, the algorithm halts and returns the last improved configuration $A^{t-1}$.

Figure~\ref{fig:method}, shows the illustration of the SFSA using a simplified example with two modalities and four potential fusion positions ($F_1$, $F_2$, $F_3$, $F_4$).
Each fusion module is visualized as a toggle switch, where red indicates that the module is inactive and green indicates that it is active.
The process begins with all fusion modules inactive, establishing a baseline configuration and its corresponding validation loss.
Next, each fusion module is tested individually, reusing the baseline’s weights to ensure efficient comparisons.
Activating module $F_3$ alone results in the lowest validation loss among the single-module configurations, as shown in the plot of the loss functions, thereby setting a new performance benchmark. Notably, the configuration with $F_2$ active also showed an improvement compared to the preceding setup, although it did not outperform the configuration with $F_3$ active.
From this best-performing single-module configuration, the algorithm then attempts to add a second module at $F_1$, $F_2$, or $F_4$.
No second module provides additional performance gains, so the algorithm selects $F_3$ alone as the optimal fusion configuration.

Determining the optimal fusion configuration by brute force would require evaluating $2^m-1$ subsets if there are $m = |\mathcal{C}|$ candidate fusion positions.
On the other hand, the proposed methodology is significantly more efficient as it incrementally explores at most $m$ additional configurations per selected fusion module, halting early when no further improvements are detected.
If the algorithm converges after selecting $r$ fusion modules, the total number of trained configurations is on the order of $r \cdot m$, which is smaller than $2^m-1$.
Furthermore, the algorithm reuses previously learned weights, reducing the cost of retraining each new configuration from scratch.
In practice, this approach allows for efficient discovery of a high-performing fusion architecture within a fraction of the computational time required by brute-force methods. The result is a practical, scalable, and data-driven strategy to determine \textit{When} to fuse modalities in multimodal deep learning, enabling improved performance in medical imaging tasks without prohibitive computational expense.

\section{Experimental Setup} \label{sec:Experimental_Setup}

In this section, we detail the datasets, preprocessing steps, training protocols, and evaluation strategies employed to validate the proposed SFSA.
We then describe the baseline unimodal and competitor models used to assess the performance of the proposed methodology.

\subsection{Datasets}
Two independent publicly available multimodal MRI datasets were employed to validate our approach.
The first, denoted as Epilepsy, acquired at the University Hospital Bonn, comprised 170 subjects, including 85 controls and 85 patients diagnosed with focal cortical dysplasia (FCD) type II~\cite{dataset1_epilessy}.
Each subject underwent 3D-T1 weighted and 3D-T2 FLAIR MRI scans, providing complementary anatomical and pathological contrasts.
This dataset aims to enhance the development and validation of automated lesion detection algorithms, particularly for FCDs that may not be easily identified through conventional MRI analysis.
The second, denoted as OASIS-3, sourced from the Washington University Knight Alzheimer Center, included 847 participants: 508 healthy controls and 339 individuals with Alzheimer’s disease~\cite{OASIS}.
In this second dataset, T1-weighted and T2-weighted MRI scans were chosen to maximize cohort size and enable effective multimodal analysis.
OASIS-3 serves as a valuable resource for researchers investigating the progression of Alzheimer's disease and the processes associated with normal aging.

All images underwent a standardized preprocessing pipeline designed to ensure consistency and data quality.
First, images from various scanners and formats, e.g., DICOM, NIfTI, were harmonized by converting all DICOM data into NIfTI format.
The resulting images were then spatially normalized by resizing and aligning them based on the most common pixel spacing values within each dataset.
To isolate the brain and remove non-relevant structures, a U-Net-based skull-stripping~\cite{bib:deepbrain2024} procedure was applied.
Finally, pixel intensities were clipped to modality-specific ranges and normalized to the $[0, 1]$ interval using a min-max scaler.

To improve model generalization, we employed spatial data augmentation during training.
Specifically, random shifts ($\pm3$ pixels) and horizontal reflections (along the $x$-axis) were applied to each modality’s images.
Identical preprocessing and augmentation steps were applied to training, validation, and test sets to ensure fair comparisons.
Both datasets were trained for a binary classification task, with the goal of distinguishing between patients and controls.

\subsection{Model and Training Configuration}
Our multimodal model integrates multiple 3D convolutional neural networks (CNNs), each based on a 3D ResNet-18 backbone~\cite{bib:hara3dcnns} pretrained on MED3D~\cite{bib:chen2019med3d}.
This architecture, comprising 18 convolutional layers organized into four residual blocks, effectively captures hierarchical spatial and structural features.
After each residual block, a MMTM~\cite{bib:MMTM} may be inserted to adaptively highlight salient features from each modality and fuse them into a shared representation.

For optimization, we adopted a supervised classification framework, using cross-entropy loss and the Adam optimizer with an initial learning rate of $10^{-4}$.
Training proceeded in minibatches of size 8 for a maximum of 300 epochs.
Early stopping was employed, halting training if validation loss did not improve for 50 consecutive epochs.
A stratified 10-fold cross-validation scheme ensured robust performance estimates, with 7 folds for training, 2 for validation, and 1 for testing.
This strategy maintained class balance across splits, providing stable and reliable evaluation.

\subsection{Baselines and Competitors}
We compared our SFSA against several baselines and competitor models to contextualize its performance and assess its relative advantages:

\paragraph{Brute-force} We considered an exhaustive configuration baseline approach that trains and evaluates all possible fusion configurations independently.
With four candidate fusion positions, this results in $2^4 - 1 = 15$ distinct multimodal configurations (excluding the configuration with no modules), and we present the performance of the best-performing configuration.
Each configuration was trained using the same preprocessing and optimization pipelines.
Comparing our method against this exhaustive baseline investigates the computational and performance benefits of our incremental, data-driven search strategy.

\paragraph{Late Fusion} Instead of integrating modalities at intermediate layers, this competitor processes each modality through separate ResNet-18 streams and fuses their predictions only at the final output stage.
While simpler, this approach does not benefit from joint feature learning at intermediate layers.
Comparing against late fusion investigates the importance of intermediate multimodal interactions and tests the necessity of systematic timing optimization.

\paragraph{Unimodal} We also included unimodal CNNs trained on individual modalities.
These models, identical in architecture to the multimodal streams, establish a lower-bound performance benchmark.
Improvements in multimodal fusion would demonstrate the added value of integrating multiple modalities and would justify the need to introduce the search of when the fusions should occur.

\section{Results and Discussions} \label{sec:Results_and_Discussions}

\begin{table*}[t]
\caption{Performance metrics on the Epilepsy dataset (mean ± standard error).}
\label{tab:epilepsy_table}
\centering
\resizebox{\textwidth}{!}{%
\begin{tabular}{lccccccc}
\toprule
\textbf{Model} & \textbf{AUC} & \textbf{Accuracy} & \textbf{F-score} & \textbf{Precision} & \textbf{Recall} & \textbf{Specificity}\\
\midrule
\textbf{SFSA} & \textbf{87.92 ± 2.96} & \textbf{81.76 ± 2.55} & \textbf{83.51 ± 2.57} & \textbf{76.59 ± 2.32} & \textbf{92.92 ± 4.11} & \textbf{70.00 ± 4.45}\\
Brute-force & 87.36 ± 2.76 & 80.59 ± 2.79 & 82.19 ± 3.23 & 75.66 ± 2.44 & 91.46 ± 4.98 & 69.51 ± 4.33\\
Late Fusion & 84.39 ± 2.54 & 81.17 ± 2.61 & 82.67 ± 3.03 & 75.51 ± 2.34 & 92.63 ± 4.51 & 68.88 ± 3.95\\
Unimodal (T1-weighted) & 81.53 ± 2.97 & 71.18 ± 2.69 & 71.52 ± 3.08 & 70.68 ± 2.79 & 74.17 ± 4.80 & 67.92 ± 4.08\\
Unimodal (T2-FLAIR) & 74.45 ± 3.51 & 78.24 ± 3.05 & 79.55 ± 4.04 & 73.00 ± 2.50 & 89.17 ± 6.02 & 66.39 ± 4.39\\
\bottomrule
\end{tabular}%
}
\end{table*}

\begin{table*}[t]
\caption{Performance metrics on the OASIS-3 dataset (mean ± standard error).}
\label{tab:OASIS3_table}
\centering
\resizebox{\textwidth}{!}{%
\begin{tabular}{lccccccc}
\toprule
\textbf{Model} & \textbf{AUC} & \textbf{Accuracy} & \textbf{F-score} & \textbf{Precision} & \textbf{Recall} & \textbf{Specificity}\\
\midrule
\textbf{SFSA} & \textbf{80.62 ± 0.32} & \textbf{74.64 ± 1.63} & \textbf{80.86 ± 0.90} & \textbf{76.11 ± 1.21} & \textbf{84.24 ± 1.13} & \textbf{60.29 ± 2.45} \\
Brute-force & 79.23 ± 1.73 & 74.47 ± 1.52 & 79.77 ± 1.28 & 75.92 ± 1.28 & 84.33 ± 1.94 & 59.74 ± 2.74 \\
Late Fusion & 75.72 ± 1.39 & 74.46 ± 1.03 & 80.60 ± 0.78 & 73.94 ± 0.97 & 89.78 ± 1.53 & 50.90 ± 2.61 \\
Unimodal (T1-weighted) & 79.55 ± 1.77 & 73.70 ± 1.30 & 79.95 ± 1.02 & 73.71 ± 1.21 & 85.62 ± 1.75 & 52.94 ± 2.87 \\
Unimodal (T2-weighted) & 70.60 ± 2.17 & 65.44 ± 1.44 & 74.87 ± 0.76 & 67.02 ± 1.68 & 85.86 ± 2.52 & 35.00 ± 3.17 \\
\bottomrule
\end{tabular}
}
\end{table*}

Tables~\ref{tab:epilepsy_table} and~\ref{tab:OASIS3_table} present the performance comparison of different models on the Epilepsy and OASIS-3 datasets, respectively, across various evaluation metrics, with the mean and standard error reported for each metric.
In both Tables~\ref{tab:epilepsy_table} and~\ref{tab:OASIS3_table}, we note that our algorithm achieves gains across all considered metrics. In each scenario, the algorithm converged on a configuration featuring a single active fusion module ($F_1$ for the Epilepsy dataset and $F_2$ for the OASIS-3 dataset), consistently outperforming unimodal baselines and other multimodal strategies.
These findings underscore the importance of identifying the optimal fusion point to effectively harness complementary information from multiple MRI sequences.
Notably, while the late fusion competitor shows some improvements over unimodal approaches, it generally fails to capture the nuanced cross-modal interactions that emerge at intermediate representation levels.
In contrast, our method’s carefully selected intermediate fusion configuration successfully integrates multimodal cues, thereby outperforming both late fusion and unimodal models and offering a more robust, data-driven approach to multimodal integration.

In addition to enhancing classification performance, the SFSA, compared to the brute-force approach, offers significant computational advantages. By incrementally introducing and evaluating fusion modules rather than exhaustively testing every potential combination from scratch, the algorithm avoids the combinatorial explosion of training costs.
This efficiency, combined with improved performance to unimodal and late-fusion competitors, supports the conclusion that optimizing fusion timing is both beneficial and tractable.

The improved performance achieved by our SFSA has important implications for clinical workflows.
By selectively fusing MRI modalities (e.g., T1-weighted and T2-FLAIR scans in epilepsy, or T1-weighted and T2-weighted scans in Alzheimer’s disease), our method harnesses complementary information that can lead to more sensitive and specific disease characterization.
Enhanced classification accuracy and more robust predictive metrics can translate into earlier diagnosis, more personalized treatment planning, and increased clinician confidence in model-driven insights.
As multimodal imaging becomes more prevalent in clinical practice, methods that efficiently identify optimal fusion strategies, like the one proposed here, could accelerate the integration of AI-driven diagnostics into routine healthcare settings.

The principles underlying this approach extend beyond medical imaging.
Any domain that relies on combining heterogeneous data sources—such as integrating structured electronic health records with genomic or wearable sensor data, or fusing multiple sensor modalities in autonomous systems, could benefit from the concepts presented here.
By systematically identifying when to fuse various data streams, this framework may inspire new methodologies in multimodal deep learning, promoting more efficient and effective architecture search in complex, high-dimensional application domains.

\section{Conclusions} \label{sec:Conclusions}

We have presented a SFSA that systematically identifies the optimal timing for modality fusion within multimodal deep learning architectures for medical imaging.
By incrementally activating and evaluating candidate fusion modules, our approach efficiently explores the architectural search space without the prohibitive computational cost of exhaustive methods.
Applied to MRI-based classification tasks, the proposed framework outperformed unimodal and late-fusion baselines, as well as brute-force combinations of multiple fusion points, yielding superior performance metrics while reducing training overhead.

These findings underscore the critical importance of fusion timing and demonstrate the utility of a data-driven, targeted methodology for modality integration.
By pinpointing where and when to fuse imaging data, our method offers a practical and scalable solution to the challenges of multimodal learning in clinical contexts.
Beyond advancing the state-of-the-art in medical image analysis, this work establishes a foundation for more adaptive, intelligent architecture design, with the potential to influence a broad range of applications spanning healthcare diagnostics and beyond.

Despite the strong results, certain limitations remain.
Our approach currently depends on a predefined set of candidate fusion points, potentially overlooking other beneficial integration stages.
Additionally, while more efficient than an exhaustive search, the SFSA still involves multiple rounds of retraining, which could be challenging in large-scale or time-sensitive clinical scenarios.
Future work may focus on adaptive strategies to propose new fusion points or pruning candidate sets based on model feedback.
Investigating advanced optimization techniques, such as gradient-based neural architecture search, or integrating model compression and acceleration could further streamline the process.
Additionally, exploring task-specific losses, domain adaptation, and transfer learning might improve generalization to diverse imaging protocols and patient populations.

\section*{Acknowledgments}
Resources are provided by the National Academic Infrastructure for Supercomputing in Sweden (NAISS) and the Swedish National Infrastructure for Computing (SNIC) at Alvis @ C3SE, partially funded by the Swedish Research Council through grant agreements no. 2022-06725 and no. 2018-05973.
This work was partially founded by: i) PNRR MUR project PE0000013-FAIR, ii) Kempe Foundation project JCSMK24-0094, iii) Università Campus Bio-Medico di Roma under the program ``University Strategic Projects'' within the project ``AI-powered Digital Twin for next-generation lung cancEr cAre (IDEA)''.

\bibliographystyle{IEEEtran}
\bibliography{mybibfile}

\end{document}